\documentclass[11pt,a4paper]{article}
\usepackage[hyperref]{emnlp-ijcnlp-2019}
\usepackage{times}
\usepackage{latexsym}
\usepackage[shortcuts]{extdash}
\usepackage{url}
\usepackage[T1]{fontenc}
\usepackage{amssymb}
\usepackage{amsmath}
\usepackage{bbm}
\usepackage{breqn}
\usepackage{algorithm}
\usepackage{algpseudocode}
\usepackage{mathptmx}
\usepackage{multicol}
\usepackage{tikz}
\usepackage{booktabs}
\usepackage{tabularx}
\usepackage{dcolumn}
\usepackage{float}
\newcolumntype{d}[1]{D{.}{.}{#1}}
\usetikzlibrary{shapes.geometric}
\usetikzlibrary{arrows.meta}

\DeclareMathAlphabet{\mathcal}{OMS}{cmsy}{m}{n}
\DeclareSymbolFont{matha}{OML}{txmi}{m}{it}
\DeclareMathSymbol{\varv}{\mathord}{matha}{118}

\aclfinalcopy 

\title{How Contextual are Contextualized Word Representations? \\ Comparing the Geometry of BERT, ELMo, and GPT-2 Embeddings}

\author{Kawin Ethayarajh\thanks{Work partly done at the University of Toronto.} \\
  Stanford University\\
  {\tt kawin@stanford.edu} \\
}

\date{}

\begin{document}
\maketitle
\begin{abstract}
  Replacing static word embeddings with contextualized word representations has yielded significant improvements on many NLP tasks. However, just how contextual are the contextualized representations produced by models such as ELMo and BERT? Are there infinitely many context-specific representations for each word, or are words essentially assigned one of a finite number of word-sense representations? For one, we find that the contextualized representations of all words are not isotropic in any layer of the contextualizing model. While representations of the same word in different contexts still have a greater cosine similarity than those of two different words, this self-similarity is much lower in upper layers. This suggests that upper layers of contextualizing models produce more context-specific representations, much like how upper layers of LSTMs produce more task-specific representations. In all layers of ELMo, BERT, and GPT-2, on average, less than 5\% of the variance in a word's contextualized representations can be explained by a static embedding for that word, providing some justification for the success of contextualized representations.
\end{abstract}

\section{Introduction}

The application of deep learning methods to NLP is made possible by representing words as vectors in a low-dimensional continuous space. Traditionally, these word embeddings were \emph{static}: each word had a single vector, regardless of context  \citep{mikolov2013distributed,pennington2014glove}. This posed several problems, most notably that all senses of a polysemous word had to share the same representation. More recent work, namely deep neural language models such as ELMo \cite{peters2018deep} and BERT \cite{devlin2018bert}, have successfully created \emph{contextualized word representations}, word vectors that are sensitive to the context in which they appear. Replacing static embeddings with contextualized representations has yielded significant improvements on a diverse array of NLP tasks, ranging from question-answering to coreference resolution.

The success of contextualized word representations suggests that despite being trained with only a language modelling task, they learn highly transferable and task-agnostic properties of language. In fact, linear probing models trained on frozen contextualized representations can predict linguistic properties of words (e.g., part-of-speech tags) almost as well as state-of-the-art models \citep{liu-gardner-belinkov-peters-smith:2019:NAACL,hewitt2019structural}. Still, these representations remain poorly understood. For one, just how contextual are these contextualized word representations? Are there infinitely many context-specific representations that BERT and ELMo can assign to each word, or are words essentially assigned one of a finite number of word-sense representations?

We answer this question by studying the geometry of the representation space for each layer of ELMo, BERT, and GPT-2. Our analysis yields some surprising findings:
\begin{enumerate}
    \item In all layers of all three models, the contextualized word representations of all words are not isotropic: they are not uniformly distributed with respect to direction. Instead, they are \emph{anisotropic}, occupying a narrow cone in the vector space. The anisotropy in GPT-2's last layer is so extreme that two random words will on average have almost perfect cosine similarity! Given that isotropy has both theoretical and empirical benefits for static embeddings \citep{mu2017all}, the extent of anisotropy in contextualized representations is surprising.
    \item Occurrences of the same word in different contexts have non-identical vector representations. Where vector similarity is defined as cosine similarity, these representations are more dissimilar to each other in upper layers. This suggests that, much like how upper layers of LSTMs produce more task-specific representations \citep{liu-gardner-belinkov-peters-smith:2019:NAACL}, upper layers of contextualizing models produce more context-specific representations. 
    \item Context-specificity manifests very differently in ELMo, BERT, and GPT-2. In ELMo, representations of words in the same sentence grow more similar to each other as context-specificity increases in upper layers; in BERT, they become more dissimilar to each other in upper layers but are still more similar than randomly sampled words are on average; in GPT-2, however, words in the same sentence are no more similar to each other than two randomly chosen words.
    \item After adjusting for the effect of anisotropy, on average, less than 5\% of the variance in a word's contextualized representations can be explained by their first principal component. This holds across all layers of all models. This suggests that contextualized representations do not correspond to a finite number of word-sense representations, and even in the best possible scenario, static embeddings would be a poor replacement for contextualized ones. Still, static embeddings created by taking the first principal component of a word's contextualized representations outperform GloVe and FastText embeddings on many word vector benchmarks.
\end{enumerate}
These insights help justify why the use of contextualized representations has led to such significant improvements on many NLP tasks. 

\section{Related Work}

\paragraph{Static Word Embeddings} Skip-gram with negative sampling (SGNS) \cite{mikolov2013distributed} and GloVe \cite{pennington2014glove} are among the best known models for generating static word embeddings. Though they learn embeddings iteratively in practice, it has been proven that in theory, they both implicitly factorize a word-context matrix containing a co-occurrence statistic \cite{levy2014linguistic,levy2014neural}. Because they create a single representation for each word, a notable problem with static word embeddings is that all senses of a polysemous word must share a single vector.

\paragraph{Contextualized Word Representations} Given the limitations of static word embeddings, recent work has tried to create context-sensitive word representations. ELMo \cite{peters2018deep}, BERT \citep{devlin2018bert}, and GPT-2 \citep{radford2019language} are deep neural language models that are fine-tuned to create models for a wide range of downstream NLP tasks. Their internal representations of words are called \emph{contextualized word representations} because they are a function of the entire input sentence. The success of this approach suggests that these representations capture highly transferable and task-agnostic properties of language \cite{liu-gardner-belinkov-peters-smith:2019:NAACL}.

ELMo creates contextualized representations of each token by concatenating the internal states of a 2-layer biLSTM trained on a bidirectional language modelling task \citep{peters2018deep}. In contrast, BERT and GPT-2 are bi-directional and uni-directional transformer-based language models respectively. Each transformer layer of 12-layer BERT (base, cased) and 12-layer GPT-2 creates a contextualized representation of each token by attending to different parts of the input sentence \cite{devlin2018bert,radford2019language}. BERT -- and subsequent iterations on BERT \citep{liu2019roberta,yang2019xlnet} -- have achieved state-of-the-art performance on various downstream NLP tasks, ranging from question-answering to sentiment analysis.

\paragraph{Probing Tasks} Prior analysis of contextualized word representations has largely been restricted to probing tasks \citep{tenney,hewitt2019structural}. This involves training linear models to predict syntactic (e.g., part-of-speech tag) and semantic (e.g., word relation) properties of words. Probing models are based on the premise that if a simple linear model can be trained to accurately predict a linguistic property, then the representations implicitly encode this information to begin with. While these analyses have found that contextualized representations encode semantic and syntactic information, they cannot answer \emph{how contextual} these representations are, and to what extent they can be replaced with static word embeddings, if at all. Our work in this paper is thus markedly different from most dissections of contextualized representations. It is more similar to \citet{mimno2017strange}, which studied the geometry of static word embedding spaces.

\section{Approach}

\subsection{Contextualizing Models} The contextualizing models we study in this paper are ELMo, BERT, and GPT-2\footnote{We use the pretrained models provided in an earlier version of the \href{https://github.com/huggingface/pytorch-transformers}{PyTorch-Transformers library}.}. We choose the base cased version of BERT because it is most comparable to GPT-2 with respect to number of layers and dimensionality. The models we work with are all pre-trained on their respective language modelling tasks. Although ELMo, BERT, and GPT-2 have 2, 12, and 12 hidden layers respectively, we also include the input layer of each contextualizing model as its $0^{\text{th}}$ layer. This is because the $0^{\text{th}}$ layer is not contextualized, making it a useful baseline against which to compare the contextualization done by subsequent layers. 

\subsection{Data} To analyze contextualized word representations, we need input sentences to feed into our pre-trained models. Our input data come from the SemEval Semantic Textual Similarity tasks from years 2012 - 2016 \cite{agirre2012semeval,agirre2013sem,agirre2014semeval,agirre2015semeval}. We use these datasets because they contain sentences in which the same words appear in different contexts. For example, the word `dog' appears in \emph{``A panda dog is running on the road.''} and \emph{``A dog is trying to get bacon off his back.''} If a model generated the same representation for `dog' in both these sentences, we could infer that there was no contextualization; conversely, if the two representations were different, we could infer that they were contextualized to some extent. Using these datasets, we map words to the list of sentences they appear in and their index within these sentences. We do not consider words that appear in less than 5 unique contexts in our analysis.

\subsection{Measures of Contextuality} We measure how contextual a word representation is using three different metrics: \emph{self-similarity}, \emph{intra-sentence similarity}, and \emph{maximum explainable variance}.

\paragraph{Definition 1} Let $w$ be a word that appears in sentences $\{ s_1, ..., s_n \}$ at indices $\{ i_1, ..., i_n \}$ respectively, such that $w = s_1[i_1] = ... = s_n[i_n]$. Let $f_{\ell}(s,i)$ be a function that maps $s[i]$ to its representation in layer $\ell$ of model $f$. The \emph{self similarity} of $w$ in layer $\ell$ is 
\begin{equation}
    \textit{SelfSim}_\ell(w) = \frac{1}{n^2 - n} \sum_{j} \sum_{k \not= j} \cos( f_{\ell}(s_j,i_j), f_{\ell}(s_k,i_k) )
\end{equation}
where $\cos$ denotes the cosine similarity. 
In other words, the \emph{self-similarity} of a word $w$ in layer $\ell$ is the average cosine similarity between its contextualized representations across its $n$ unique contexts. If layer $\ell$ does not contextualize the representations at all, then $\textit{SelfSim}_\ell(w) = 1$ (i.e., the representations are identical across all contexts). The more contextualized the representations are for $w$, the lower we would expect its self-similarity to be.

\paragraph{Definition 2} Let $s$ be a sentence that is a sequence $\left< w_1, ..., w_n \right>$ of $n$ words. Let $f_{\ell}(s,i)$ be a function that maps $s[i]$ to its representation in layer $\ell$ of model $f$. The \emph{intra-sentence similarity} of $s$ in layer $\ell$ is
\begin{equation}
\begin{split}
    \textit{IntraSim}_\ell(s) &= \frac{1}{n} \sum_{i} \cos(\vec{s_{\ell}}, f_{\ell}(s, i)) \\
    \text{where } \vec{s_{\ell}} &= \frac{1}{n} \sum_i f_{\ell}(s, i) \\
\end{split}
\end{equation}
Put more simply, the \emph{intra-sentence similarity} of a sentence is the average cosine similarity between its word representations and the sentence vector, which is just the mean of those word vectors. This measure captures how context-specificity manifests in the vector space. For example, if both $\textit{IntraSim}_\ell(s)$ and $\textit{SelfSim}_\ell(w)$ are low $\forall\ w \in s$, then the model contextualizes words in that layer by giving each one a context-specific representation that is still distinct from all other word representations in the sentence. If $\textit{IntraSim}_\ell(s)$ is high but $\textit{SelfSim}_\ell(w)$ is low, this suggests a less nuanced contextualization, where words in a sentence are contextualized simply by making their representations converge in vector space.

\paragraph{Definition 3} Let $w$ be a word that appears in sentences $\{ s_1, ..., s_n \}$ at indices $\{ i_1, ..., i_n \}$ respectively, such that $w = s_1[i_1] = ... = s_n[i_n]$. Let $f_{\ell}(s,i)$ be a function that maps $s[i]$ to its representation in layer $\ell$ of model $f$. Where $[ f_{\ell}(s_1, i_1) ... f_{\ell}(s_n, i_n) ]$ is the \emph{occurrence matrix} of $w$ and $\sigma_1 ... \sigma_m$ are the first $m$ singular values of this matrix, the \emph{maximum explainable variance} is 
\begin{equation}
    \textit{MEV}_\ell(w) = \frac{\sigma_1^2}{\sum_i \sigma_i^2}
\end{equation}
$\textit{MEV}_\ell(w)$ is the proportion of variance in $w$'s contextualized representations for a given layer that can be explained by their first principal component. It gives us an upper bound on how well a static embedding could replace a word's contextualized representations. The closer $\textit{MEV}_\ell(w)$ is to 0, the poorer a replacement a static embedding would be; if $\textit{MEV}_\ell(w) = 1$, then a static embedding would be a perfect replacement for the contextualized representations.

\subsection{Adjusting for Anisotropy}\label{anisotropy_define}

It is important to consider isotropy (or the lack thereof) when discussing contextuality. For example, if word vectors were perfectly isotropic (i.e., directionally uniform), then $\textit{SelfSim}_\ell(w) = 0.95$ would suggest that $w$'s representations were poorly contextualized. However, consider the scenario where word vectors are so anisotropic that any two words have on average a cosine similarity of 0.99. Then $\textit{SelfSim}_\ell(w) = 0.95$ would actually suggest the opposite -- that $w$'s representations were well contextualized. This is because representations of $w$ in different contexts would on average be more dissimilar to each other than two randomly chosen words.

To adjust for the effect of anisotropy, we use three \emph{anisotropic baselines}, one for each of our contextuality measures. For self-similarity and intra-sentence similarity, the baseline is the average cosine similarity between the representations of uniformly randomly sampled words from different contexts. The more anisotropic the word representations are in a given layer, the closer this baseline is to 1. For maximum explainable variance (MEV), the baseline is the proportion of variance in uniformly randomly sampled word representations that is explained by their first principal component. The more anisotropic the representations in a given layer, the closer this baseline is to 1: even for a random assortment of words, the principal component would be able to explain a large proportion of the variance. 

Since contextuality measures are calculated for each layer of a contextualizing model, we calculate separate baselines for each layer as well. We then subtract from each measure its respective baseline to get the \emph{anisotropy-adjusted contexuality measure}. For example, the anisotropy-adjusted self-similarity is
\begin{equation}
\begin{split}
    \textit{Baseline}(f_{\ell}) &= \mathbb{E}_{x,y \sim U(\mathcal{O})} \left[ \cos( f_{\ell}(x), f_{\ell}(y) ) \right]\\
    \textit{SelfSim}_\ell^*(w) &= \textit{SelfSim}_\ell(w) - \textit{Baseline}(f_{\ell}) \\
\end{split}
\end{equation}
where $\mathcal{O}$ is the set of all word occurrences and $f_{\ell}(\cdot)$ maps a word occurrence to its representation in layer $\ell$ of model $f$. Unless otherwise stated, references to contextuality measures in the rest of the paper refer to the anisotropy-adjusted measures, where both the raw measure and baseline are estimated with 1K uniformly randomly sampled word representations.

\begin{figure*}[t]
    \centering
    \includegraphics[scale=0.64]{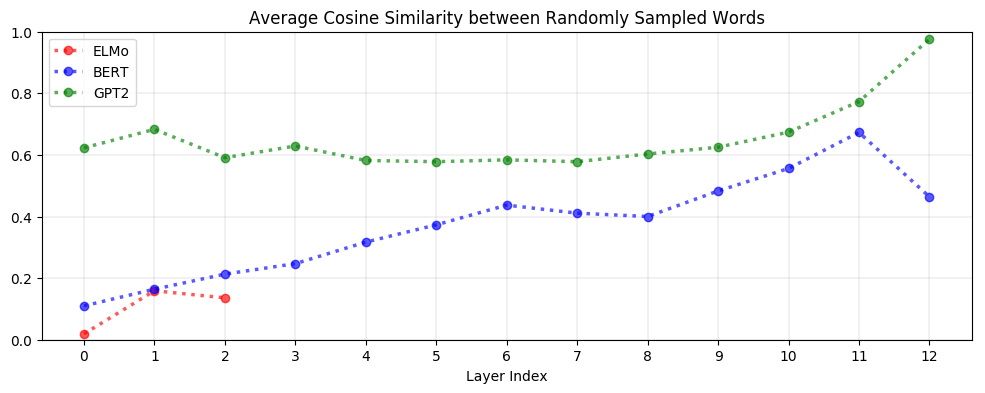}
    \caption{In almost all layers of BERT, ELMo, and GPT-2, the word representations are anisotropic (i.e., not directionally uniform): the average cosine similarity between uniformly randomly sampled words is non-zero. The one exception is ELMo's input layer; this is not surprising given that it generates character-level embeddings without using context. Representations in higher layers are generally more anisotropic than those in lower ones. }
    \label{fig:mean_cosine_similarity_across_words}
\end{figure*}

\section{Findings}

\subsection{(An)Isotropy}\label{anisotropy}

\paragraph{Contextualized representations are anisotropic in all non-input layers.} If word representations from a particular layer were isotropic (i.e., directionally uniform), then the average cosine similarity between uniformly randomly sampled words would be 0 \cite{arora2016simple}. The closer this average is to 1, the more anisotropic the representations. The geometric interpretation of anisotropy is that the word representations all occupy a narrow cone in the vector space rather than being uniform in all directions; the greater the anisotropy, the narrower this cone \citep{mimno2017strange}. As seen in Figure \ref{fig:mean_cosine_similarity_across_words}, this implies that in almost all layers of BERT, ELMo and GPT-2, the representations of all words occupy a narrow cone in the vector space. The only exception is ELMo's input layer, which produces static character-level embeddings without using contextual or even positional information \cite{peters2018deep}. It should be noted that not all static embeddings are necessarily isotropic, however; \citet{mimno2017strange} found that skipgram embeddings, which are also static, are not isotropic.

\paragraph{Contextualized representations are generally more anisotropic in higher layers.} As seen in Figure \ref{fig:mean_cosine_similarity_across_words}, for GPT-2, the average cosine similarity between uniformly randomly words is roughly 0.6 in layers 2 through 8 but increases exponentially from layers 8 through 12. In fact, word representations in GPT-2's last layer are so anisotropic that any two words have on average an almost perfect cosine similarity! This pattern holds for BERT and ELMo as well, though there are exceptions: for example, the anisotropy in BERT's penultimate layer is much higher than in its final layer.

Isotropy has both theoretical and empirical benefits for static word embeddings. In theory, it allows for stronger ``self-normalization'' during training \citep{arora2016simple}, and in practice, subtracting the mean vector from static embeddings leads to improvements on several downstream NLP tasks \cite{mu2017all}. Thus the extreme degree of anisotropy seen in contextualized word representations -- particularly in higher layers -- is surprising. As seen in Figure \ref{fig:mean_cosine_similarity_across_words}, for all three models, the contextualized hidden layer representations are almost all more anisotropic than the input layer representations, which do not incorporate context. This suggests that high anisotropy is inherent to, or least a by-product of, the process of contextualization.

\begin{figure*}[t]
    \centering
    \includegraphics[scale=0.64]{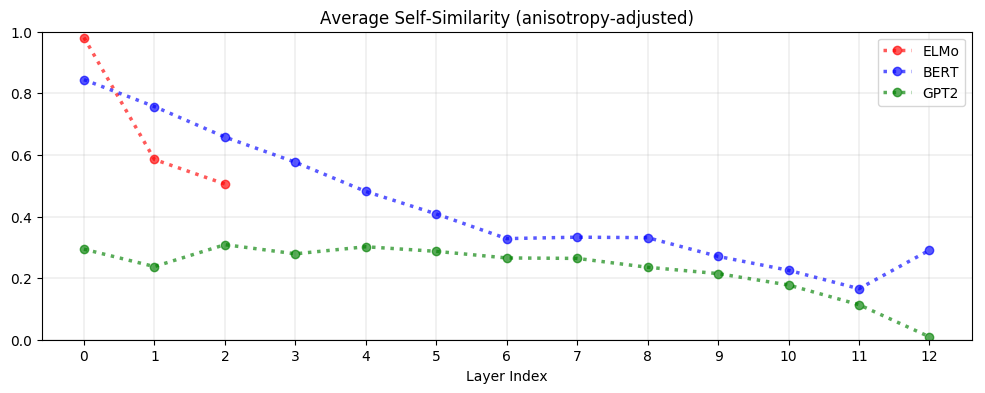}
    \caption{The average cosine similarity between representations of the same word in different contexts is called the word's \emph{self-similarity} (see Definition 1). Above, we plot the average self-similarity of uniformly randomly sampled words after adjusting for anisotropy (see section \ref{anisotropy_define}). In all three models, the higher the layer, the lower the self-similarity, suggesting that contextualized word representations are more context-specific in higher layers.}
    \label{fig:self_similarity_above_expected}
\end{figure*}

\begin{figure*}[t]
    \centering
    \includegraphics[scale=0.64]{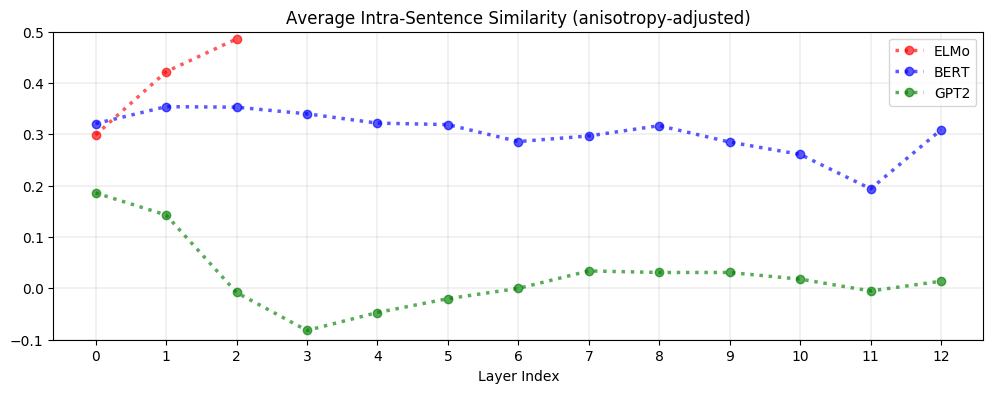}
    \caption{The \emph{intra-sentence similarity} is the average cosine similarity between each word representation in a sentence and their mean (see Definition 2). Above, we plot the average intra-sentence similarity of uniformly randomly sampled sentences, adjusted for anisotropy. This statistic reflects how context-specificity manifests in the representation space, and as seen above, it manifests very differently for ELMo, BERT, and GPT-2.}
    \label{fig:mean_cosine_similarity_between_sentence_and_words}
\end{figure*}

\subsection{Context-Specificity}

\paragraph{Contextualized word representations are more context-specific in higher layers.} Recall from Definition 1 that the self-similarity of a word, in a given layer of a given model, is the average cosine similarity between its representations in different contexts, adjusted for anisotropy. If the self-similarity is 1, then the representations are not context-specific at all; if the self-similarity is 0, that the representations are maximally context-specific. In Figure \ref{fig:self_similarity_above_expected}, we plot the average self-similarity of uniformly randomly sampled words in each layer of BERT, ELMo, and GPT-2. For example, the self-similarity is 1.0 in ELMo's input layer because representations in that layer are static character-level embeddings.

In all three models, the higher the layer, the lower the self-similarity is on average. In other words, the higher the layer, the more context-specific the contextualized representations. This finding makes intuitive sense. In image classification models, lower layers recognize more generic features such as edges while upper layers recognize more class-specific features \cite{yosinski2014transferable}. Similarly, upper layers of LSTMs trained on NLP tasks learn more task-specific representations \cite{liu-gardner-belinkov-peters-smith:2019:NAACL}. Therefore, it follows that upper layers of neural language models learn more context-specific representations, so as to predict the next word for a given context more accurately. Of all three models, representations in GPT-2 are the most context-specific, with those in GPT-2's last layer being almost maximally context-specific.

\paragraph{Stopwords (e.g., \emph{`the', `of', `to'}) have among the most context-specific representations.} Across all layers, stopwords have among the lowest self-similarity of all words, implying that their contextualized representations are among the most context-specific. For example, the words with the lowest average self-similarity across ELMo's layers are \emph{`and', `of', `'s', `the'}, and \emph{`to'}. This is relatively surprising, given that these words are not polysemous. This finding suggests that the variety of contexts a word appears in, rather than its inherent polysemy, is what drives variation in its contextualized representations. This answers one of the questions we posed in the introduction: ELMo, BERT, and GPT-2 are not simply assigning one of a finite number of word-sense representations to each word; otherwise, there would not be so much variation in the representations of words with so few word senses.

\begin{figure*}[t]
    \centering
    \includegraphics[scale=0.64]{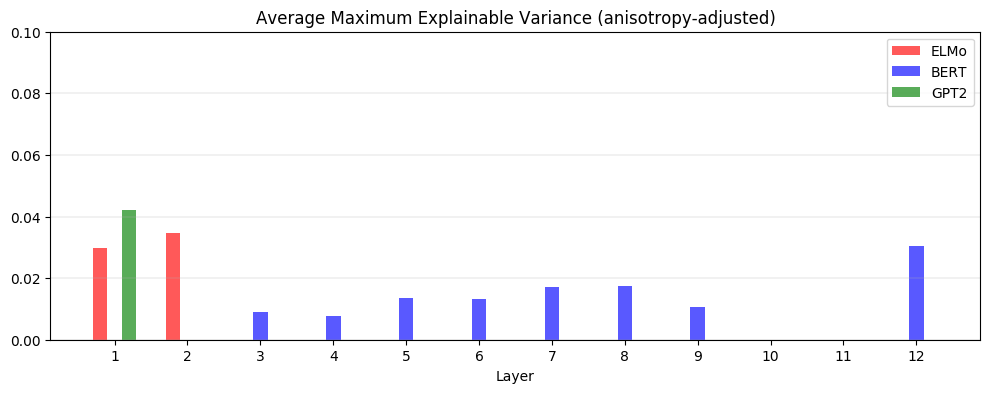}
    \caption{The \emph{maximum explainable variance} (MEV) of a word is the proportion of variance in its contextualized representations that can be explained by their first principal component (see Definition 3). Above, we plot the average MEV of uniformly randomly sampled words after adjusting for anisotropy. In no layer of any model can more than 5\% of the variance in a word's contextualized representations be explained by a static embedding.}
    \label{fig:variance_explained}
\end{figure*}

\paragraph{Context-specificity manifests very differently in ELMo, BERT, and GPT-2.} As noted earlier, contextualized representations are more context-specific in upper layers of ELMo, BERT, and GPT-2. However, how does this increased context-specificity manifest in the vector space? Do word representations in the same sentence converge to a single point, or do they remain distinct from one another while still being distinct from their representations in other contexts? To answer this question, we can measure a sentence's intra-sentence similarity. Recall from Definition 2 that the intra-sentence similarity of a sentence, in a given layer of a given model, is the average cosine similarity between each of its word representations and their mean, adjusted for anisotropy. In Figure \ref{fig:mean_cosine_similarity_between_sentence_and_words}, we plot the average intra-sentence similarity of 500 uniformly randomly sampled sentences. 

\paragraph{In ELMo, words in the same sentence are more similar to one another in upper layers.} As word representations in a sentence become more context-specific in upper layers, the intra-sentence similarity also rises. This suggests that, in practice, ELMo ends up extending the intuition behind Firth's \citeyearpar{firth1957synopsis} distributional hypothesis  to the sentence level: that because words in the same sentence share the same context, their contextualized representations should also be similar.

\paragraph{In BERT, words in the same sentence are more dissimilar to one another in upper layers.} As word representations in a sentence become more context-specific in upper layers, they drift away from one another, although there are exceptions (see layer 12 in Figure \ref{fig:mean_cosine_similarity_between_sentence_and_words}). However, in all layers, the average similarity between words in the same sentence is still greater than the average similarity between randomly chosen words (i.e., the anisotropy baseline). This suggests a more nuanced contextualization than in ELMo, with BERT recognizing that although the surrounding sentence informs a word's meaning, two words in the same sentence do not necessarily have a similar meaning because they share the same context.

\paragraph{In GPT-2, word representations in the same sentence are no more similar to each other than randomly sampled words.} On average, the unadjusted intra-sentence similarity is roughly the same as the anisotropic baseline, so as seen in Figure \ref{fig:mean_cosine_similarity_between_sentence_and_words}, the anisotropy-adjusted intra-sentence similarity is close to 0 in most layers of GPT-2. In fact, the intra-sentence similarity is highest in the input layer, which does not contextualize words at all. This is in contrast to ELMo and BERT, where the average intra-sentence similarity is above 0.20 for all but one layer. 

As noted earlier when discussing BERT, this behavior still makes intuitive sense: two words in the same sentence do not necessarily have a similar meaning simply because they share the same context. The success of GPT-2 suggests that unlike anisotropy, which accompanies context-specificity in all three models, a high intra-sentence similarity is not inherent to contextualization. Words in the same sentence can have highly contextualized representations without those representations being any more similar to each other than two random word representations. It is unclear, however, whether these differences in intra-sentence similarity can be traced back to differences in model architecture; we leave this question as future work.
 
\begin{table*}[ht!]
    \centering 
    \footnotesize
    \begin{tabularx}{\textwidth}{X ccccccccc}
 
 \toprule
Static Embedding &  SimLex999 &    MEN &  WS353 &     RW &  Google &    MSR &  SemEval2012(2) &  BLESS &     AP \\
\midrule
GloVe &      0.194 &  0.216 &  0.339 &  0.127 &   0.189 &  0.312 &          0.097 &  0.390 &  0.308 \\
FastText        &      0.239 &  \bf 0.239 &  \bf 0.432 &  0.176 &   0.203 &  0.289 &          0.104 &  0.375 &  0.291 \\
ELMo, Layer 1         &      0.276 &  0.167 &  0.317 &  0.148 &   0.170 &  0.326 &          0.114 &  \bf 0.410 &  0.308 \\
ELMo, Layer 2       &      0.215 &  0.151 &  0.272 &  0.133 &   0.130 &  0.268 &          0.132 &  0.395 &  0.318 \\
BERT, Layer 1         &      0.315 &  0.200 &  0.394 & \bf  0.208 &   \bf 0.236 & \bf 0.389 & \bf 0.166 &  0.365 &  \bf 0.321 \\
BERT, Layer 2         &      \bf 0.320 &  0.166 &  0.383 &  0.188 &   0.230 &  0.385 &          0.149 &  0.365 & \bf 0.321 \\
BERT, Layer 11       &      0.221 &  0.076 &  0.319 &  0.135 &   0.175 &  0.290 &          0.149 &  0.370 &  0.289 \\
BERT, Layer 12        &      0.233 &  0.082 &  0.325 &  0.144 &   0.184 &  0.307 &          0.144 &  0.360 &  0.294 \\
GPT-2, Layer 1         &      0.174 &  0.012 &  0.176 &  0.183 &   0.052 &  0.081 &          0.033 &  0.220 &  0.184 \\
GPT-2, Layer 2         &      0.135 &  0.036 &  0.171 &  0.180 &   0.045 &  0.062 &          0.021 &  0.245 &  0.184 \\
GPT-2, Layer 11        &      0.126 &  0.034 &  0.165 &  0.182 &   0.031 &  0.038 &          0.045 &  0.270 &  0.189 \\
GPT-2, Layer 12        &      0.140 & -0.009 &  0.113 &  0.163 &   0.020 &  0.021 &          0.014 &  0.225 &  0.172 \\
\bottomrule

    \end{tabularx}
    \caption{The performance of various static embeddings on word embedding benchmark tasks. The best result for each task is in bold. For the contextualizing models (ELMo, BERT, GPT-2), we use the first principal component of a word's contextualized representations in a given layer as its static embedding. The static embeddings created using ELMo and BERT's contextualized representations often outperform GloVe and FastText vectors.   }
    \label{tab:benchmark}
\end{table*}

\subsection{Static vs.\ Contextualized}

\paragraph{On average, less than 5\% of the variance in a word's contextualized representations can be explained by a static embedding.} Recall from Definition 3 that the \emph{maximum explainable variance} (MEV) of a word, for a given layer of a given model, is the proportion of variance in its contextualized representations that can be explained by their first principal component. This gives us an upper bound on how well a static embedding could replace a word's contextualized representations. Because contextualized representations are anisotropic (see section \ref{anisotropy}), much of the variation \emph{across all words} can be explained by a single vector. We adjust for anisotropy by calculating the proportion of variance explained by the first principal component of uniformly randomly sampled word representations and subtracting this proportion from the raw MEV. In Figure \ref{fig:variance_explained}, we plot the average anisotropy-adjusted MEV across uniformly randomly sampled words.

In no layer of ELMo, BERT, or GPT-2 can more than 5\% of the variance in a word's contextualized representations be explained by a static embedding, on average. Though not visible in Figure \ref{fig:variance_explained}, the raw MEV of many words is actually below the anisotropy baseline: i.e., a greater proportion of the variance across all words can be explained by a single vector than can the variance across all representations of a single word. Note that the 5\% threshold represents the best-case scenario, and there is no theoretical guarantee that a word vector obtained using GloVe, for example, would be similar to the static embedding that maximizes MEV. This suggests that contextualizing models are not simply assigning one of a finite number of word-sense representations to each word -- otherwise, the proportion of variance explained would be much higher. Even the average raw MEV is below 5\% for all layers of ELMo and BERT; only for GPT-2 is the raw MEV non-negligible, being around 30\% on average for layers 2 to 11 due to extremely high anisotropy.

\paragraph{Principal components of contextualized representations in lower layers outperform GloVe and FastText on many benchmarks.} As noted earlier, we can create static embeddings for each word by taking the first principal component (PC) of its contextualized representations in a given layer. In Table \ref{tab:benchmark}, we plot the performance of these \emph{PC static embeddings} on several benchmark tasks\footnote{The \href{https://github.com/kudkudak/word-embeddings-benchmarks}{Word Embeddings Benchmarks} package was used for evaluation.}. These tasks cover semantic similarity, analogy solving, and concept categorization: SimLex999 \cite{hill2015simlex}, MEN \cite{bruni2014multimodal}, WS353 \cite{finkelstein2002placing}, RW \cite{luong2013better}, SemEval-2012 \cite{jurgens2012semeval}, Google analogy solving \cite{mikolov2013distributed} MSR analogy solving  \cite{mikolov2013linguistic}, BLESS \cite{baroni2011we} and AP \cite{almuhareb2004attribute}. 
We leave out layers 3 - 10 in Table \ref{tab:benchmark} because their performance is between those of Layers 2 and 11.

The best-performing PC static embeddings belong to the first layer of BERT, although those from the other layers of BERT and ELMo also outperform GloVe and FastText on most benchmarks.  For all three contextualizing models, PC static embeddings created from lower layers are more effective those created from upper layers. Those created using GPT-2 also perform markedly worse than their counterparts from ELMo and BERT. Given that upper layers are much more context-specific than lower layers, and given that GPT-2's representations are more context-specific than ELMo and BERT's (see Figure \ref{fig:self_similarity_above_expected}), this suggests that the PCs of highly context-specific representations are less effective on traditional benchmarks. Those derived from less context-specific representations, such as those from Layer 1 of BERT, are much more effective.

\section{Future Work}

Our findings offer some new directions for future work. For one, as noted earlier in the paper, \citet{mu2017all} found that making static embeddings more isotropic -- by subtracting their mean from each embedding -- leads to surprisingly large improvements in performance on downstream tasks. Given that isotropy has benefits for static embeddings, it may also have benefits for contextualized word representations, although the latter have already yielded significant improvements despite being highly anisotropic. Therefore, adding an anisotropy penalty to the language modelling objective -- to encourage the contextualized representations to be more isotropic -- may yield even better results. 

Another direction for future work is generating static word representations from contextualized ones. While the latter offer superior performance, there are often challenges to deploying large models such as BERT in production, both with respect to memory and run-time. In contrast, static representations are much easier to deploy. Our work in section 4.3 suggests that not only it is possible to extract static representations from contextualizing models, but that these extracted vectors often perform much better on a diverse array of tasks compared to traditional static embeddings such as GloVe and FastText. This may be a means of extracting some use from contextualizing models without incurring the full cost of using them in production.

\section{Conclusion}

In this paper, we investigated how contextual contextualized word representations truly are. For one, we found that upper layers of ELMo, BERT, and GPT-2 produce more context-specific representations than lower layers. This increased context-specificity is always accompanied by increased anisotropy. However, context-specificity also manifests differently across the three models; the anisotropy-adjusted similarity between words in the same sentence is highest in ELMo but almost non-existent in GPT-2. We ultimately found that after adjusting for anisotropy, on average, less than 5\% of the variance in a word's contextualized representations could be explained by a static embedding. This means that even in the best-case scenario, in all layers of all models, static word embeddings would be a poor replacement for contextualized ones. These insights help explain some of the remarkable success that contextualized representations have had on a diverse array of NLP tasks.

\section*{Acknowledgments}

We thank the anonymous reviewers for their insightful comments. We thank the Natural Sciences and Engineering Research Council of Canada (NSERC) for their financial support.

\bibliography{acl2018}
\bibliographystyle{acl_natbib}
\end{document}